\documentclass[runningheads]{llncs}
\usepackage{amsmath}

\usepackage{multirow}

\usepackage[T1]{fontenc}
\usepackage{graphicx}
\usepackage{hyperref}
\usepackage{color}

\usepackage{amssymb}
\usepackage{booktabs}

\DeclareMathOperator*{\argmax}{arg\,max}
\DeclareMathOperator*{\argmin}{arg\,min}

\begin{document}
\title{Investigating the Interplay between Features and Structures in Graph Learning}
\titlerunning{The Interplay between Features and Structures in Graph Learning}
%
\author{Daniele Castellana\inst{1} \and
Federico Errica \inst{2}
}
%
\authorrunning{D. Castellana, F. Errica}
%
\institute{University of Florence, Florence, Italy \\\email{daniele.castellana@unifi.it} \\ \and
NEC Laboratories Europe, Heidelberg, Germany \\
\email{federico.errica@neclab.eu}}
%
\maketitle              
\begin{abstract}
In the past, the dichotomy between homophily and heterophily has inspired research contributions toward a better understanding of Deep Graph Networks' inductive bias. In particular, it was believed that homophily strongly correlates with better node classification predictions of message-passing methods. More recently, however, researchers pointed out that such dichotomy is too simplistic as we can construct node classification tasks where graphs are completely heterophilic but the performances remain high. Most of these works have also proposed new quantitative metrics to understand when a graph structure is useful, which implicitly or explicitly assume the correlation between node features and target labels. Our work empirically investigates what happens when this strong assumption does not hold, by formalising two generative processes for node classification tasks that allow us to build and study ad-hoc problems. To quantitatively measure the influence of the node features on the target labels, we also use a metric we call Feature Informativeness. We construct six synthetic tasks and evaluate the performance of six models, including structure-agnostic ones. Our findings reveal that previously defined metrics are not adequate when we relax the above assumption. Our contribution to the workshop aims at presenting novel research findings that could help advance our understanding of the field.

\end{abstract}

\section{Introduction}
\label{sec:introduction}
In recent years we have witnessed an exponentially increasing interest in machine learning techniques that can process graph-structured data \cite{bronstein_geometric_2017,hamilton_representation_2017,wu_comprehensive_2020,bacciu_gentle_2020}. This is due to the wide range of applications that these methods can deal with, as graphs are a natural abstraction in many scientific fields: chemistry \cite{reiser_graph_2022}, physics \cite{carleo_machine_2019}, and recommender systems \cite{fan_graph_2019} to name a few. 

In the context of Deep Graph Networks (DGNs) \cite{bacciu_gentle_2020}, based on neural \cite{scarselli_graph_2009,micheli_neural_2009}, probabilistic \cite{bacciu_probabilistic_2020}, or even untrained message passing \cite{gilmer_neural_2017}, researchers have empirically observed that such models perform favourably on node classification tasks when the graph structure is homophilic \cite{mcpherson_birds_2001}, meaning that adjacent nodes in the graph share similar features or target labels to be predicted. In contrast, heterophilic graphs exhibit an opposite behaviour, and structure-agnostic baselines like a Multi-Layer Perceptron (MLP) proved to be better or very competitive compared to DGNs at classifying nodes under heterophily \cite{zhu_beyond_2020,maekawa_beyond_2022}. There has been a lot of effort in the literature to mitigate the (apparent) detrimental effects of heterophily, but a separate line of works recently showed that it is possible to construct completely heterophilic graphs where DGNs achieve an almost perfect test performance \cite{ma_homophily_2022,cavallo_2_2023}\footnote{We found some mistakes in the proofs of this paper, which however do not undermine its empirical results. We contacted the authors who are working on a new version.}. There, it is claimed that the distribution of the neighbouring class labels is a decisive factor to determine whether a DGN's inductive bias is suitable for the task, meaning that it leads to better performances compared to an MLP. These results, however, rely on a specific, and often implicit, assumption about a good correlation between the node features and the target.

The goal of this contribution is to continue the discussion on how to determine what are the components that make a node classification task difficult to solve, without being necessarily bound to message-passing architectures. At present time, we know that homophily alone is not a good metric to gauge the difficulty of the task, and in this paper we aim at showing that if we relax the assumptions of \cite{ma_homophily_2022}, the distribution of the neighbouring class labels does not tell the whole story either. We achieve this by evaluating different models on a few but carefully designed synthetic datasets for node classification that control specific properties of the graph, similarly to what is done in \cite{palowitch_graphworld_2022}. In particular, we give examples of cases where an unfavourable distribution of neighbouring class labels can be associated with very good DGNs performances.

Our observations open new questions that we would like to discuss in the context of the workshop: is it possible to create a taxonomy of the characteristics that make a task easy or hard to solve (e.g., for message-passing architectures)? Can we find new measures to quantitatively assess the presence or absence of such characteristics? Is it possible to find criteria to determine the intrinsic difficulty of tasks when the assumptions of \cite{ma_homophily_2022} do not hold? 

Addressing these questions can be critical for many reasons. First of all, it would enhance our understanding of DGNs' inductive bias (which is related but orthogonal to their expressiveness in distinguishing substructures \cite{inverno_2021_new}), thus allowing us to make sensible choices of model architectures for specific tasks. Second, it could give some practical guidelines to identify whether the structure of the graph gives a practical advantage compared to just using the node features, which (as of today) remains a purely empirical question \cite{errica_fair_2020}. In turn, this would allow people to successfully apply DGNs to problems where no structure is available, by creating ad-hoc structures with well-defined characteristics. So far, this has been done with nearest-neighbour graphs, although it remains unclear whether this brings a significant advantage compared to an MLP \cite{li_spam_2019,benamira_semi_2019,malone_learning_2021,ragesh_hetegcn_2021,yu_resgnet_2021,lu_nagnn_2022,tong_predicting_2022}.

\section{Related Works}
\label{sec:related-works}
The body of literature that critically investigates the apparent harm of heterophilic graphs has recently been growing. Some authors have tried to define different ``node behaviours'' by analyzing the degree of a node and its local level of heterophily, arguing that they can explain whether a DGN will perform well or badly on a task \cite{yan_two_2022}. These results rely on the assumption that nodes from different classes exhibit distinct (to some extent) distributions of their features. This is precisely the assumption we want to challenge in our contribution because, when put to the extreme, it means a structure-agnostic baseline would already be able to correctly separate all nodes. Another contribution introduced a new metric that measures how different node embeddings of different classes look after an aggregation step, showing a good correlation between high values of the metric and good performances of a DGN \cite{luan_revisiting_2022}. Others exploited the homophily assumption on the node features to rewire the graph by connecting homophilic but distant neighbours, under the assumption that homophily helps DGNs to perform better \cite{li_finding_2022}.

In the following, we will build on some recently defined metrics such as Label Informativeness (LI) \cite{platonov_characterizing_2022} and Cross-Class Neighborhood Similarity (CCNS) \cite{ma_homophily_2022} to create synthetic tasks. These are metrics that look more at the class information rather than the features, and so far they have shown to correlate well with the performance of a DGN when features of nodes of different classes are generated by distinct (i.e. well-enough separated) distributions.

In time, researchers have also tried to empirically assess the quality of DGNs compared to structure-agnostic baselines on a plethora of different real and synthetic tasks \cite{shchur_pitfalls_2018,errica_fair_2020,dwivedi_benchmarking_2020,hu_open_2020,palowitch_graphworld_2022}. Their results are mostly quantitative, but they might act as a ``certificate of goodness'' of newly developed measures of tasks' difficulty. Recently, an attempt has been made in this direction by comparing the performance of models across a range of node and graph properties \cite{maekawa_beyond_2022}, but some of the conclusions do not explicitly mention all assumptions made in the process; for example, it appears that DGNs works better in the homophilic setup, but we know from other works that this is not necessarily the case.  

Finally, most theoretical research on message-passing methods deals with the issues of over-smoothing \cite{li_deeper_2018,chen_measuring_2020,chamberlain_grand_2021,bodnar_neural_2022}, which is concerned with the flattening of the graph signal after repeated applications of graph convolutions, and over-squashing \cite{alon_bottleneck_2021,topping_understanding_2022,gravina_anti_2023} of learned representations, meaning that there is a bottleneck of information when predictions have to be made. In addition, there is a very active research direction on the expressive power that comes with each message-passing model \cite{morris_weisfeiler_2019,xu_how_2019,geerts_let_2021,bodnar_weisfeiler_2021}. These works are orthogonal to the current discussion, as they talk about how difficult it can be for DGNs to process particular graphs regardless of the specific task under consideration.

Overall, the landscape of contributions is vast and it is easy to misinterpret some of the results when important assumptions are not explicitly formalized. In the next sections, we will consider the scenario where the assumption of the correlation between node features and target labels is not valid.

\section{Method}
\label{sec:method}
In this section, we formalise a generative process to create synthetic node classification tasks, each defined by a graph. We will focus on undirected graphs $G=(\mathcal{V},\mathcal{E})$, where $\mathcal{V}$ and $\mathcal{E}$ represent the set of nodes and edges, respectively. The edges in $\mathcal{E}$ define the structure of $G$: two nodes $u\in\mathcal{V}$ and $v \in \mathcal{V}$ are connected, or adjacent, if an edge $\{u,v\}$ belongs to $\mathcal{E}$. The set of neighbours of node $u \in \mathcal{V}$ is denoted by $\mathcal{N}_u$, that is, $\mathcal{N}_u = \{v \in \mathcal{V} \mid \{u,v\} \in \mathcal{E}\}$; also, the degree of a node is defined as $d_u = |\mathcal{N}_u|$. In this paper, each node has a feature vector and a target class label attached to it. For a node $u$, we denote its input feature and output label as $x_u$ and $y_u$, respectively. Without loss of generality, each input feature is a $d$-dimensional vector, meaning $x_u \in \mathbb{R}^d$, and each class label $y_u$ belongs to a finite set of labels $\mathcal{C}$.

Before introducing the new generative process that we use in our experiments, we review the most common metrics used in the literature to quantify homophily, heterophily, and other relevant properties in graphs. A subsequent analysis of their limitations will motivate our systematic study of different node classification tasks by explicitly defining their generative processes and thus the underlying assumptions.

\subsubsection{Homophily/Heterophily Quantities}
Quantitative metrics for measuring homophily in graphs rely on the class similarity between adjacent node pairs. In particular, a graph is considered homophilic if most of the adjacent nodes share the same class label. In contrast, in an heterophilic graph nodes of the same class tend not to connect together. 
The most common definition is possibly the ``edge-homophily'' \cite{zhu_beyond_2020}:
\begin{equation}
    h_{e} = \frac{\sum_{\{u,v\} \in \mathcal{E}} \mathbb{I}[y_u = y_v]}{|\mathcal{E}|},
\end{equation}
where $\mathbb{I}[pred]=1$ if and only if the predicate $pred$ is true. A purely heterophilic graph has $h_{e}=0$, whereas a purely homophilic graph satisfies $h_{e}=1$.

A related quantity is the node-homophily \cite{Pei2020Geom-GCN:}, which computes the average proportion of neighbours that have the same class across all nodes in the graph. These two quantities are simple and intuitive, but they are sensitive to the number of classes and their imbalance, making the numbers hard to interpret and not very comparable across different datasets \cite{lim2021large,platonov_characterizing_2022}. To overcome this issue, researchers have introduced the notion of ``adjusted homophily'' \cite{platonov_characterizing_2022}:
\begin{equation}
    h_{adj} = \frac{h_e - \sum_{k\in \mathcal{C}}\bar{p}(k)^2}{1-\sum_{k\in\mathcal{C}}\bar{p}(k)^2},
\end{equation}
where $\bar{p}(k) \propto \sum_{u \in \mathcal{V}_k} d_u$, and $\mathcal{V}_k = \{u \in \mathcal{V} \mid y_u = k \}$ is the subset of nodes with the node class label equal to $k$. 
We will consider both edge and adjusted homophily in our experimental settings to see how different they can get. 

\subsubsection{Recent Quantities}
More recently, researchers realised that the dichotomy between homophily and heterophily does not tell the whole story about how ``good'' a graph structure is, where good means that said structure allows to better separate nodes of different classes with the right inductive bias \cite{ma_homophily_2022,platonov_critical_2023,platonov_characterizing_2022}. Consequently, they started focusing on capturing neighbouring patterns across nodes of the same or different classes.

The Cross-Class Neighbourhood Similarity (CCNS) \cite{ma_homophily_2022} computes the similarity between two classes $k,k'$ using the neighbourhood class label distribution:
\begin{equation}
    \text{CCNS}(k, k') = \frac{1}{|\mathcal{V}_k||\mathcal{V}_{k'}|} \sum_{u\in\mathcal{V}_k}\sum_{v \in \mathcal{V}_{k'}} \text{cos}(D_u, D_v),
\end{equation}
where $D_u$ is the empirical histogram (over $|\mathcal{C}|$ classes) of node $u$’s neighbouring class labels, that is, $D_u(t) = \sum_{w\in\mathcal{N}_u} \mathbb{I}[y_w=t]$ counts how many neighbours of $u$ have the class label $t$. The function $\text{cos}(\cdot,\cdot)$ computes the cosine similarity between two histograms, but any other similarity function can be used.
Importantly, the elements on the diagonal measure the neighbourhood similarity among nodes with the same class label; instead, the off-diagonal elements measure the neighbourhood similarity among nodes with different class labels. The CCNS has empirically proven to be a good certificate of difficulty for a task: when there is high intra-class similarity and low inter-class similarity, a Graph Convolutional Network (GCN) \cite{kipf_semi-supervised_2017} typically performs better compared to an MLP \cite{ma_homophily_2022}.

The Label Informativeness (LI) \cite{platonov_characterizing_2022} is another metric that measures the impact of neighbours' class labels to determine a particular node's class label:
\begin{equation}
    \text{LI} = 2 - \frac{\sum_{k \in \mathcal{C}}\sum_{k' \in \mathcal{C}}p(k,k')\log p(k,k')}{\sum_{k \in \mathcal{C}} \bar{p}(k) \log \bar{p}(k)},
\end{equation}
where $p(k,k') \propto \sum_{\{u,v\} \in \mathcal{E}} \mathbb{I}[y_u = k \land  y_v=k']$ is the empirical joint distribution of class labels along the edges, and $\bar{p}(k) \propto \sum_{u \in \mathcal{V}_k} d_u$. Given an edge $\{u,v\} \in \mathcal{E}$, LI measures the amount of knowledge the label $y_v$ gives for predicting $y_u$. Its value is equal to 0 when the node label does not depend on the neighbours' labels, and it is equal to 1 when the node label is completely determined by one of its neighbours' labels.
Similarly to the CCNS, the LI empirically correlates with the performances of DGNs better than other homophily metrics \cite{platonov_characterizing_2022}.

\subsubsection{Limitations of Current Approaches}

The CCNS and LI metrics seem to give more insights into what is a helpful graph structure, in the sense that the numbers they provide correlate with the performance of a graph machine learning classifier. However, these encouraging results rely on a strong assumption: the input features $x_u$ are generated from a distribution $F_{y_u}$, and the expectation of the different distributions $F_k$ are assumed to be distinct from each other \cite{ma_homophily_2022}. While this assumption might seem reasonable (especially for tabular datasets), it is also too simplistic for at least two reasons. First, it assumes that there is already a good degree of separability between the nodes by just looking at their feature values. Secondly, we can build tasks where the node label depends solely on the structural property of the input graph and not on their feature values; in this case, the node features might even be missing. Generally speaking, a task can be solved even though the relationship between node features and target labels is too complex to show a direct correlation.

\subsection{Generative Processes for Graphs}
As mentioned in Section \ref{sec:introduction}, our investigation stems from a few simple questions: what happens when the assumption on features we just described does not hold anymore? Can we still use quantitative metrics such as the CCNS and LI to determine how ``good'' a structure is? How badly can these quantities fail at recognising a helpful structure? 
To investigate this interplay between features and structures in graph learning, we introduce two generative processes to create synthetic node classification tasks with relatively straightforward characteristics. In both cases, we will be able to enforce a low correlation between the features and the target, which we measure with a quantity called Feature Informativeness (FI) (also known as the coefficient of determination in previous works \cite{malvestuto_statistical_1986}).
The idea of FI is similar to LI: given that we model the input features $\mathbf{x}$ and the class labels $\mathbf{y}$ as random variables, we measure the amount of knowledge the feature $\mathbf{x}$ gives for predicting $\mathbf{y}$ as:
\begin{equation}
    \text{FI} = \frac{I(\mathbf{x},\mathbf{y})}{H(\mathbf{x})},
\end{equation}
where $I(\cdot, \cdot)$ is the (empirical) mutual information and $H(\cdot)$ is the (empirical) entropy, with $0 \le I(\mathbf{x},\mathbf{y}) \le H(\mathbf{x})$. When the input features are not categorical, we discretise them using a histogram with a bin size equal to 1. The value of FI ranges in the interval $[0,1]$, where $0$ (resp. $1$) denotes minimal (resp. maximal) informativeness between node features and target labels.

\subsubsection{Neighbours-Based Generative Process.}
The first generative process builds tasks where the class label of a node depends solely on its neighbours:
\begin{equation}
\begin{aligned}
    h_u & \sim \text{Cat}(1/K,\dots,1/K), &
    x_u \mid \mathbf{h} &\sim \mathcal{N}(\mu_{h_u},\sigma_{h_u}), &
    \mathcal{E} \mid \mathbf{h} &\sim E(\mathbf{h}), \\
    H_{uk}&=\sum_{v\in\mathcal{N}_u}\mathbb{I}[h_v = k], & y_u \mid \mathcal{E}, \mathbf{h} &= \phi(H_{u1},\dots,H_{uK}) & &
\end{aligned}
\end{equation}
where all $h_u$ are hidden factors modelled as categorical variables with $K$ states and the node input feature $x_u$ is generated from a Gaussian distribution whose parameters depend on $h_u$. The node class label $y_u$ is generated using a deterministic function $\phi$ which takes as input $K$ values $H_{u1},\dots, H_{uK}$, where $H_{uk}$ counts the number of neighbours whose hidden variables are in the $k$-th state. 
Thanks to the introduction of the hidden variables and the choice of $\phi$, we can highlight failure cases of the known metrics and argue for the creation of new ones that are effective when the value of FI is low.

\subsubsection{Structural-Based Generative Process.}
In the second generative process, the class label of a node depends solely on some structural properties and not necessarily on neighbouring information. We define such process as:
\begin{equation}
    \begin{aligned}
    y_u & \sim \text{Cat}(1/K,\dots,1/K), &
    \mathcal{E} \mid \mathbf{y} &\sim E(\mathbf{y}), &
    x_u \mid \mathcal{E} &= d_u,\\
    \end{aligned}
\end{equation}
where all $y_u$ are categorical variables with $K$ states and the node input feature $x_u$ is the node degree. The distribution $E$ generates a structure whose edges are dependent on the specific node label assignments $\mathbf{y}$; we refer the reader to the next section for concrete examples. By construction, the node features 
are not required at all to solve the task. Here, we use the node degree as the only node feature similar to previous works \cite{errica_fair_2020} to obtain a low FI score.

\section{Experiments}
\label{sec:experiments}
In the following, we report the details to reproduce our results, from the task generation to the training of the different baselines.\footnote{\url{https://github.com/danielecastellana22/feature-structure-interplay-graph-learning}}

\subsection{Task Generation}

We create six different node classification tasks: the first three are generated by the neighbour-based generative process (symbol $N_i$) while the last three are generated by the structural one (symbol $S_i$). 

\subsubsection{Most Common Hidden State ($N_1$)}
In the first neighbour-based task $N_1$, we fix $K=4$ and generate a random structure (i.e. for each node we randomly select its neighbours). A continuous node feature is generated using 4 different Gaussian, one for each class.
The function $\phi$ is defined as follows:
\begin{equation}
    y_u = \phi_1(H_{u1},H_{u2},H_{u3},H_{u4}) = \argmax_{k \in [1,4]} H_{uk}.
    \label{eq:most_common}
\end{equation}

\subsubsection{Least Common Hidden State ($N_2$)}
The second task $N_2$ is obtained by just replacing the $\argmax$ of Equation \eqref{eq:most_common} with the $\argmin$. To simplify the task from a ``structural'' viewpoint, we ensure that, for each node, it exists a hidden state that never occurs in its neighbourhood (i.e. $\forall u\in\mathcal{V}.\, \exists k \in [1,4].\, H_k=0$).

\subsubsection{Parity ($N_3$)}
The last neighbour-based task uses yet another different $\phi$ that generates the class node label as follows:
\begin{equation}
    \phi_3(H_{u1}, H_{u2}) = H_{u2} \bmod 2.
\end{equation}
This means we consider the parity of the number of neighbours of class 2 to determine the class of each node. In $N_3$, the number of hidden states (and thus the number of node class labels) is equal to $2$.

\subsubsection{Easy Multipartite ($S_1$)} 
The first task purely based on structural properties is task $S_1$, where we generate a multipartite graph with $K=4$ clusters. The label $y_u$ indicates the cluster of a node. To facilitate the task, the connectivity pattern among clusters is further simplified: nodes in cluster $1$ are connected with nodes in cluster $4$ and nodes in cluster $2$ are linked to nodes in cluster $3$. Thus, each node has neighbours with the same label. 

\subsubsection{Random Multipartite ($S_2$)} 
The structural task $S_2$ is again a multipartite graph with $K=4$ clusters. However, this time the structure is generated randomly. Given a node $u$ and its cluster $y_u$, we randomly select its neighbours from all the nodes outside its cluster, i.e. the set $\{v \mid y_v \neq y_u\}$. Thus, each node has neighbours with all labels different from its own. 

\subsubsection{Count Triangles ($S_3$)}
In the last structural task, the node class label $y_u$ indicates the number of triangles in which the node is involved. At first, we sample the labels $y_u \in [1,4]$ for all nodes ($K=4$). Then, we generate a structure such that, for all nodes, the value of $y_u$ effectively indicate the number of triangles the node belongs to. There exist various structures that can satisfy these constraints; in our experiment, we simply use the algorithm proposed in \cite{miller_percolation_2009,newman_random_2009} to generate a valid structure. We obtain the final structure by removing self-loops and parallel edges.

\subsection{Task Statistics}
\label{subsec:task-stats}

\setlength{\tabcolsep}{3pt}
\begin{table}[t]
    \centering
    \begin{tabular}{lcccccccc}
        \toprule
         & $|\mathcal{V}|$ & $|\mathcal{E}|$ & Avg. $d_u$ & $K=|\mathcal{C}|$ & $h_e$ & $h_{adj}$ & LI & FI\\
        \midrule
        $N_1$ & 1911 & 32418 & 16.9639 & 4 & 0.2566 & -0.0053 & 0.0003 & 0.0093\\
        $N_2$ & 1600 & 33416 & 20.8850 & 4 & 0.2475 & -0.0037 & 0.0003 & 0.0112\\
        $N_3$ & 1600 & 31790 & 19.8687 & 2 & 0.5014 & 0.0025 & 0.0000 & 0.0029\\
        \midrule
        $S_1$ & 1600 & 33524 & 20.9525 & 4 & 0.0000 & -0.3333 & 1.0000 & 0.0182\\
        $S_2$ & 1600 & 33398 & 20.8738 & 4 & 0.0000 & -0.3333 & 0.2076 & 0.0164\\
        $S_3$ & 1580 & 9288 & 5.8785 & 4 & 0.3402 & 0.1152 & 0.0464 & 0.2704\\
        \bottomrule
    \end{tabular}
    \caption{We report the statistics of the tasks considered. All our tasks exhibit a low Feature Informativeness.}
    \label{tab:data_stats}
\end{table}

In Table \ref{tab:data_stats}, we report some statistics of the generated tasks. First, we observe that the FI scores are always close to zero except for $S_3$, and all graphs appear to be quite heterophilic since the maximal values of $h_{adj}$ is 0.11.
All generated graphs exhibit different label connectivity patterns: in the neighbour-based tasks, there is no information gain in observing the label of a neighbour in isolation 
(i.e. all tasks have $\text{LI}=0$); instead, in the structure-based tasks, we have three different and higher values for LI.
\begin{figure}[t]
    \centering
    \includegraphics[width=\textwidth]{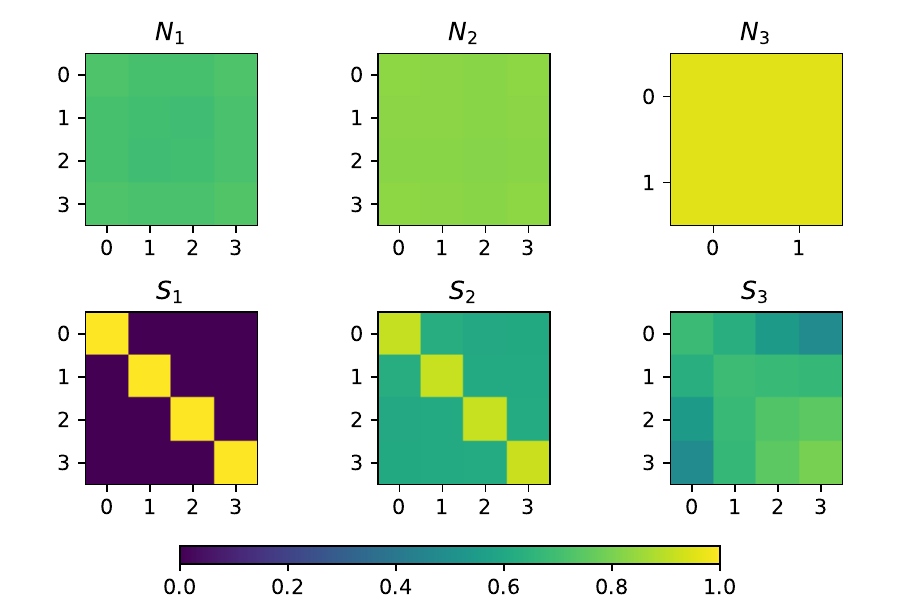}
    \caption{We show the CCNS matrices for every task considered. Our tasks exhibit a broad spectrum of CCNs matrices with low FI.}
    \label{fig:data_ccns}
\end{figure}

We also show the CCNS matrix for each task in Figure \ref{fig:data_ccns}. All neighbour-based tasks (first row) have an almost uniform CCNS matrix,
meaning that not even the neighbouring label distribution of a node is informative. 
In this case, one can easily see a relation to the very low values of LI in Table \ref{tab:data_stats}. 
Instead, in the first two structural tasks (second row) the CCNS exhibits a clear diagonal pattern, whereas in $S_3$ the pattern is less obvious; this indicates that the neighbouring label distribution of a node contains some information about its label. Interestingly, here there is less agreement between the CCNS and the LI: while on $S_1$ the value of LI is equal to 1 (i.e. the label of a node is completely determined by a neighbouring label), it is only $0.2$ on $S_2$. As pointed out in \cite{platonov_characterizing_2022}, this happens because LI only considers the effect of picking one label from the neighbourhood rather than considering the whole neighbouring label distribution (as CCNS does). As a side note, we notice that the CCNS matrix in $S_1$ is different from the one in $S_2$ even if in both tasks each node can be uniquely classified by looking at the neighbouring class label distribution. This behaviour is likely due to the limited expressiveness of the cosine similarity, which cannot capture all differences between probability distributions. 

\begin{figure}[t]
    \centering
    \includegraphics[width=\textwidth]{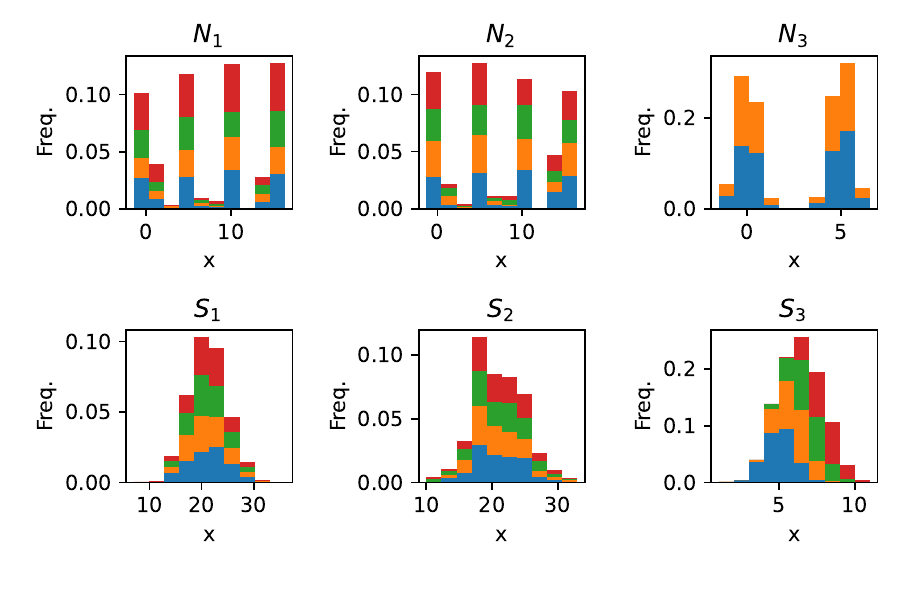}
    \caption{We visualise the (discretised) distributions of the continuous features for the different tasks, divided by class value. Except for S3, the distribution of the feature values is approximately the same for each class.}
    \label{fig:fetaure_distr_plot}
\end{figure}
For completeness, Figure \ref{fig:fetaure_distr_plot} depicts the histograms of the node feature for each class and each task. In all the neighbour-based tasks (first row) we can clearly distinguish the different Gaussians used to generate the feature, but there is almost no correlation with the target labels thanks to our generative process. \\
Similarly, in the structural tasks (second row), the target labels have been generated in a completely independent way from the node features. The higher FI for $S_3$ is merely an artefact due to our arbitrary choice of the degree as a node feature.

These analyses reveal that \textit{when the target labels do not directly correlate with the node features}, but instead depend on the underlying generative process in complex ways, metrics such as LI and CCNS suggest that DGNs might perform on par or worse than an MLP on the first three tasks ($N_1, N_2, N_3$), whereas the opposite should be true for the last three ($S_1, S_2, S_3$). In the next sections, we will empirically show that this is indeed not the case, thus questioning the effectiveness of these metrics in discriminating the ``goodness'' of the graph structure when the FI is low.

\subsection{Experimental Settings}
For each node classification task, we consider an experimental setup where we randomly split 70\% of the nodes for training, 10\% for validation, and 20\% for testing. To assess the model performances, we first carry out model selection by selecting the best configuration on the validation set, where accuracy is the metric of interest. After the best configuration is selected, we estimate the trained model's empirical risk on the test nodes. This process is repeated 10 times for different splits and the test results are averaged together. 

The models considered in our experiments are GCN \cite{kipf_semi-supervised_2017}, GATv2 \cite{brody_attentive_2022}, GraphSAGE \cite{hamilton_inductive_2017}, PNA \cite{corso_principal_2020}, and an MLP. In addition, we consider another baseline called ``Mode'' that always predicts the most frequent label in the training data. In the model selection, we tried the following hyper-parameters: latent dimension $\in \{2,5,10,20\}$, learning rate $\in \{0.01\}$, number of epochs $\in \{1000\}$, early-stopping patience $\in \{100\}$, number of layers $\in \{2,\dots,5\}$ for the neighbour-based tasks and $\in \{2,\dots,9\}$ for the structure-based ones, number of heads $\in \{1,2,3\}$ for GAT and aggregation function $\in \{\text{mean}, \text{max}\}$ for Graph-SAGE. For all DGNs, we also validated: graph augmentation with self-loops, input skip-connections and the ego- and neighbour-embedding separation (as suggested in previous works for heterophilic graphs \cite{zhu_beyond_2020}).

\subsection{Results}
\begin{table}
    \centering 
    \begin{tabular}{rcccccc}
        \toprule
         & $N_1$ & $N_2$ & $N_3$ & $S_1$ & $S_2$ & $S_3$\\
         \midrule
        GCN & $99.1 \pm 0.4$ & $72.9 \pm 1.7$ & $53.9 \pm 3.0$ & $99.5 \pm 0.5$ & $37.4 \pm 1.8$ & $54.9 \pm 0.9$\\
        GATv2 & $99.0 \pm 0.5$ & $74.6 \pm 1.5$ & $58.0 \pm 5.7$ & $99.1 \pm 0.5$ & $59.2 \pm 11.4$ & $57.0 \pm 2.0$\\
        GraphSAGE & $98.9 \pm 0.4$ & $73.3 \pm 1.4$ & $53.3 \pm 4.2$ & $99.2 \pm 0.7$ & $70.6 \pm 7.3$ & $54.9 \pm 1.5$\\
        PNA & $98.9 \pm 0.5$ & $72.9 \pm 1.0$ & $53.4 \pm 3.0$ & $99.2 \pm 0.4$ & $49.0 \pm 5.1$ & $55.4 \pm 1.2$\\
        \midrule
        MLP & $32.0 \pm 2.4$ & $26.5 \pm 1.8$ & $51.6 \pm 1.6$ & $25.7 \pm 1.7$ & $25.8 \pm 1.0$ & $49.6 \pm 1.0$\\
        Mode & $32.7 \pm 0.5$ & $26.0 \pm 0.6$ & $50.9 \pm 0.8$ & $25.0 \pm 0.6$ & $24.8 \pm 0.6$ & $26.1 \pm 0.6$\\
        \bottomrule
    \end{tabular}
    \caption{Mean and std test accuracy results obtained by models on the tasks.}
    \label{tab:exp_results}
\end{table}

In Table \ref{tab:exp_results} we report the accuracy of all models under study on our synthetic tasks. On the neighbour-based tasks ($N_1, N_2, N_3$), the MLP has significantly poorer performances compared to the DGNs because it does not have access to the important information carried by the connectivity patterns, and it has comparable accuracy to the Mode baseline. This result comes as no surprise due to the absence of a correlation between features and target labels. The DGNs, on the other hand, are able to exploit the structural information and produce better scores. Nevertheless, the synthetic tasks are associated with different degrees of difficulty: while on $N_1$ all DGNs reach a test accuracy of almost 99\%, they achieve only 74\% and 58\% on $N_2$ and $N_3$, respectively. Following up on the discussion of Section \ref{subsec:task-stats}, the CCNS and LI are in clear disagreement with the empirical results of Table \ref{tab:exp_results}: a very low value of LI and an almost uniform CCNS matrix correspond to very good DGN scores on $N_1$.

In the structure-based tasks ($S_1, S_2, S_3$), the MLP still struggles. On $S_1$ and $S_2$, its performances are close to the Mode baseline, while on $S_3$ it is able to reach a higher accuracy of almost 50\% while Mode is stuck at 26\%. This behaviour is justified by the higher value of feature informativeness reported in Table \ref{tab:data_stats}.
Also in this case, the DGNs always outperform the structure-agnostic baselines since the class labels depend on the structure, and as before the values of accuracy obtained by DGNs vary across the task. While all models solve $S_1$ by reaching an accuracy of 99\%, the results on $S_2$ range from 37\% (GCN) to 70\% (GraphSAGE). Both $S_1$ and $S_2$ are multipartite graphs with the following difference: in $S_1$ the class of a node is completely determined by a label from its neighbourhood (since all neighbours have the same label), while in $S_2$ the node class label is determined by all the neighbours' label (it is the only missing labels in the neighbours). The different performances obtained are therefore due to the intrinsic difficulties in detecting the presence of neighbours rather than their absence. In fact, in $S_1$ it is sufficient to observe one of the neighbours' labels to predict the correct node class; instead, in $S_2$ we need to know all the neighbours' labels to detect the missing class. This is yet another difference that cannot be detected by metrics like the CCNS matrix since the neighbours' label distribution is different for each class (as shown in Figure \ref{fig:data_ccns}). Finally, on $S_3$ all DGNs reach an accuracy that is slightly better than the MLP one. The hardness of this task arises from the complexity of counting the number of triangles in random graphs, and it cannot be captured by previous metrics that do not focus on the structural properties of the graph. For instance, the CCNS would seem to suggest that $S_3$ should have been simpler than $N_1$, but this is not what we have observed empirically. 

In summary, these results show that class label-based metrics alone are not enough to capture the intrinsic difficulty of neighbour-based and structural tasks when the node features do not correlate with class labels, which is the main and strong assumption that previous works have (more or less implicitly) made. 
This calls for the development of new metrics and further investigation of data where the FI is low.
We also argue that researchers have not devoted enough attention yet to the nature of the function $\phi$. While we do not have access to this function in real-world problems, deepening our understanding of the relation between $\phi$ and DGNs' performances in a controlled setting can help us make considerable progress in the field.


\section{Conclusions}
\label{sec:conclusions}
In this paper, we identified potential limitations of known node classification metrics that estimate how useful the graph structure is in effectively distinguishing nodes belonging to different classes. We then presented two generative processes for node classification tasks where the main assumption of these metrics does not hold. Through our findings, it became evident that both the CCNS and LI are inadequate indicators when node features do not correlate with target labels. This work hopes to highlight the necessity for novel indicators to fill this gap and to emphasise the importance of a more systematic investigation into the underlying assumptions of these metrics. Such efforts are crucial if we want to make concrete progress in the area of graph representation learning.

\newpage
\bibliographystyle{plain}
\bibliography{bibliography}


\end{document}